# A Hierarchal Planning Framework for AUV Mission Management in a Spatio-Temporal Varying Ocean

**Somaiyeh MahmoudZadeh, David M.W Powers, Karl Sammut, Adham Atyabi, Amir Mehdi Yazdani**

Centre for Maritime Engineering, Control and Imaging,
School of Computer Science, Engineering and Mathematics, Flinders University, Adelaide, SA, Australia

somaiyeh.mahmoudzadeh@flinders.edu.au
david.powers@flinders.edu.au
karl.sammut@flinders.edu.au
adham.atyabi@seattlechildrens.org
amirmehdi.yazdani@flinders.edu.au

**Corresponding author:**

Somaiyeh MahmoudZadeh

School of Computer Science, Engineering and Mathematics, Flinders University, Tonsley Park
SA 5042, Australia

E-mail: somaiyeh.mahmoudzadeh@flinders.edu.au

# A Hierarchal Planning Framework for AUV Mission Management in a Spatio-Temporal Varying Ocean

**Abstract-** The purpose of this paper is to provide a hierarchical dynamic mission planning framework for a single autonomous underwater vehicle (AUV) to accomplish task-assign process in a limited time interval while operating in an uncertain undersea environment, where spatio-temporal variability of the operating field is taken into account. To this end, a high level reactive mission planner and a low level motion planning system are constructed. The high level system is responsible for task priority assignment and guiding the vehicle toward a target of interest considering on-time termination of the mission. The lower layer is in charge of generating optimal trajectories based on sequence of tasks and dynamicity of operating terrain. The mission planner is able to reactively re-arrange the tasks based on mission/terrain updates while the low level planner is capable of coping unexpected changes of the terrain by correcting the old path and re-generating a new trajectory. As a result, the vehicle is able to undertake the maximum number of tasks with certain degree of maneuverability having situational awareness of the operating field. The computational engine of the mentioned framework is based on the biogeography based optimization (BBO) algorithm that is capable of providing efficient solutions. To evaluate the performance of the proposed framework, firstly, a realistic model of undersea environment is provided based on realistic map data, and then several scenarios, treated as real experiments, are designed through the simulation study. Additionally, to show the robustness and reliability of the framework, Monte-Carlo simulation is carried out and statistical analysis is performed. The results of simulations indicate the significant potential of the two-level hierarchical mission planning system in mission success and its applicability for real-time implementation.

Autonomous operation of AUVs in a vast, unfamiliar and dynamic underwater environment is a complicated process, especially when the AUV is obligated to react to environment changes, where usually a-priory information is not available. Recent advancements in sensor technology and embedded computer systems has opened new possibilities in underwater path planning and made AUVs more capable for handling complicated long range underwater missions. However, there still exist major challenges for this class of the vehicle, where the surrounding environment has a complex spatio-temporal variability and uncertainty. Ocean current variability affect vehicle's motion, for example it can perturb its safety by pushing that to an undesired direction [1]. Consequently, this variability can also have a profound impact on vehicle's battery usage and its mission duration. The robustness of a vehicle's path planning to this strong environment variability is a key element to its safety and mission performance. Thus, robustness of the trajectory planning to current variability and terrain uncertainties is essential to mission success and AUV safe deployment. On the other hand, an AUV should carry out complex tasks in a limited time interval. However, existing AUVs have limited battery capacity and restricted endurance, so they should be capable of managing mission time. Obviously, a single AUV is not able to meet all specified tasks in a single mission with limited time and energy, so the vehicle has to effectively manage its resources to perform effective persistent deployment in longer missions without human interaction. In this respect, time management is a fundamental requirement toward mission success that tightly depends on the optimality of the selected tasks between start and destination point in a graph-like operation terrain. Hereupon, design of an efficient mission planning framework considering vehicle's availabilities and capabilities is essential requirement for maximizing mission productivity. Many efforts have been devoted in recent years for enhancing an AUV's capability in robust motion planning and efficient task assignment. Although some improvement have been achieved in other autonomous systems; there are still many challenges to achieve a satisfactory level of intelligence and robustness for AUV in this regard.

AUVs capabilities in handling mission objectives are directly influenced by routing and task arrangement system performance. Effective routing has a great impact on a vehicle's time management as well as mission performance by appropriate selection and arrangement of the tasks sequence. Many attempts have been carried out in scope of single or multiple vehicles routing for different purposes. Route planning problems usually is referred as finding shortest paths in a graph-like network such as modelling the transportation network [2,3]. The main issue that should be covered by route planning system is to direct vehicle(s) to its destination in a network while providing efficient maneuver and optimizing travel time. Route planning on multi-agent decisions is implemented by Dominik et al., for the purpose of transport planning, where the agent decides between distributing orders to customers, traversing edges, competing vendors, increasing production, etc [4]. Zou et al., investigated application of Genetic Algorithm (GA) in dynamic route guidance system [5]. A behavior-based controller coupled with waypoint tracking scheme is introduced by Karimanzira et al., for AUV guidance in large-scale underwater environment [6]. An integrated mission assignment and routing strategy is proposed in [7] to serve the AUVs routing problem in order to deliver customized sensor packages to mission targets at scattered positions, while minimising total energy cost in the presence of ocean currents. The AUV routing problem is investigated with a Double Traveling Salesman Problem with Multiple Stacks (DTSPMS) for a single-vehicle pickup-and-delivery problem by minimizing the total routing cost [8]. A large scale route planning and task assignment joint problem related to the AUV activity has been investigated in [9] by transforming the problem space into a NP-hard graph context and using the heuristic search nature of Genetic Algorithm (GA) and Particle Swarm Optimization (PSO) to find the best waypoints. Later on, the same concept is extended by M.Zadeh et al., to AUV routing in a semi dynamic network, while performance of the BBO and PSO algorithm is tested on single vehicle's routing approach [10]. The real-time performance of the application is usually overshadowed by growth of the graph complexity or problem search space. Growth of the search space increases the computational burden that is often a problematic issue with deterministic methods such as mixed integer linear programming (MILP) proposed by Yilmaz et al., for governing multiple AUVs [11]. In terms of task assignment, many typical AUV missions are limited to executing a list of pre-programmed instructions and completing a predefined sequences of tasks.

The majority of the mentioned research particularly focuses on task and target assignment and time scheduling problems without considering requirements for vehicle's safe deployment or quality of its motion in presence of environmental disturbances. A vehicle's safe and confident deployment is a critical issue that should be taken into consideration at all stages of the mission in a vast and uncertain environment. In the rest of this section, existing AUV trajectory/path planning approaches are discussed, which are more concentrated on vehicles deployment encountering dynamicity of the operating environment

Various strategies have been developed and applied to the AUV path-planning problem in recent years. The well-known direct method of optimal control theory, called inverse dynamics in the virtual domain (IDVD) method, was employed to develop and test a real-time trajectory generator for realization on board of an AUV [12,13]. Willms and Yang proposed a real-time collision-free robot path planning based on a dynamic-programming shortest path algorithm [14]. A sliding wave front expansion algorithm applying continuous optimization techniques has been presented by Soulignac for AUVs path planning in presence of strong current fields [15]. Jan et al., investigated higher geometry maze routing algorithm for optimal path planning and navigating a mobile rectangular robot among obstacles [16]. Nevertheless, this strategy may not be appropriate for AUV dynamic environments where the current field changes continuously during the mission. Earlier proposed methods [14,16] are capable of providing optimum path planning for AUV using previous information for replanning process, which is computationally reasonable in generating accurate local trajectories; however, they modelled the environment as a 2D space, which is inefficient for application of the AUVs, as a 2D representation of a marine environment doesn't sufficiently embody all the information of a 3D ocean environment and the vehicle's six degree of freedom. The evolution-based strategies like Differential Evolution (DE) [17], GA [18,19] and PSO [20] are another approach that has been applied successfully to the path planning problem and are fast enough to satisfy time restrictions of the real-time applications. A real-time online evolution based path planner was developed for AUV rendezvous path planning in a cluttered variable environment, in which the performance of four evolutionary algorithms of Firefly Algorithm (FA), BBO, DE, and PSO is tested and compared in different scenarios [21]. A Quantum-based PSO (QPSO) was applied by Fu et al., for unmanned aerial vehicle's path planning, in which only the off-line path planning in a static known environment was implemented, which is not enough to cover dynamicity of the underwater environment [22]. Later on, this algorithm was employed by Zeng et al., for AUV's on-line path planning in dynamic marine environment [1].

Although various path planning techniques have been suggested for autonomous vehicles, AUV-oriented applications still have several difficulties when operating across a large-scale geographical area. The recent investigations on path planning that incorporate variability of the environment have assumed that planning is carried out with perfect knowledge of probable future changes of the environment [23,24], while in the reality, accurate prediction of the environmental events such as currents or obstacles state variations is impractical specially in longer operations. Even though available ocean predictive approaches operate reasonably well in small scales and over short time periods, they produce insufficient accuracy to current prediction over long time periods in larger scales, specifically in cases with lower information resolution [25]. Moreover, current variations over time can affect moving obstacles (or waypoints in some cases) and drift them across a vehicle's trajectory; therefore, the planned trajectory may change and become invalid or inefficient. Proper estimation of the events in such a dynamic uncertain terrain in long range operations, outside the vehicle's sensor coverage, is impractical and unreliable. This becomes even more challenging in larger dimensions, when a large data load of variation of whole terrain condition should be computed repeatedly any time that path replanting is required, which is computationally inefficient and unnecessary as only awareness of environment changes in vicinity of the vehicle is enough.

As mentioned earlier, the path planning problem principally deals with the quality of a vehicle's motion between two points and it is not an appropriate strategy to carry out vehicle's task assignment and mission timing in a graph-like terrain. On the other hand, vehicles routing strategies are not flexible like path planning methods in terms of handling environment sudden changes, but they give a general overview of the area that an AUV should fly through (general route), which means reducing the operation area to smaller operating zone for vehicle's deployment.

To satisfy the addressed challenges and to produce a reliable mission plan for a large scale time-varying underwater environments, this paper proposes a reactive hybrid framework that comprises an efficient mission planning system combined with real-time path planning that improves a vehicle's ability to complete as much of its mission as possible within the time available. Furthermore, a path planner is designed in a smaller scale to concurrently plan trajectory between waypoints included in the task sequence. The path planner operates in the context of the mission planner in a manner to be fast enough to handle unforeseen changes and regenerates an alternative trajectory that safely guides the vehicle through the specified waypoints with minimum time/energy cost. A constant interaction exists between high-low level planners across small and large scale. This paper is a continuation of previous research [26] in which the environment modeled to be more realistic comprising uncertainty of moving/afloat objects and dynamic multiple-layered time varying ocean current; accordingly, the path planner in current study is facilitated with dynamic re-planning capability, which have not been addressed in the previous study. The reactive hybrid model decides whether to carry out the path re-planning or mission re-planning procedure according to the raised situation. The path re-planning is performed to cope with dynamic changes of the operating environment over time. On the other hand, mission re-planning is performed to manage the lost time in cases that the path planner process takes longer than expectation. The proposed re-planning procedure in both of the mission and path planners improve the robustness and reactive ability of the AUV to the environmental changes and enhance its performance in accurate mission timing. Both of the planners operate individually and concurrently while sharing their information. Parallel execution of the planners speed up the computation process.

In the core of the proposed strategy, both mission planner and local path planners make use of the Biogeography-based Optimization (BBO) algorithm. The argument for application of BBO in solving Non-deterministic Polynomial-time (NP) hard problems is strong enough due to its remarkable competency in scaling with multi-objective and complex problems. In this algorithm solutions of one generation are transferred to the next and never discarded but modified. This characteristic of the BBO enhances its exploitation ability. Solving NP-hard problems is computationally challenging and currently there is no polynomial time algorithm to handle a NP-hard problem of even moderate size. Furthermore, finding a pure optimum solution is only possible when the environment is fully known and no uncertainty exists. The modeled underwater environment in this paper corresponds to a highly dynamic uncertain environment. The BBO is one of the fastest meta–heuristics algorithms introduced for solving NP-hard complex problems. Although the captured solutions do not necessarily correspond to a pure optimal solution, controlling the computational time is more preferable in this research due to the real-time application of the AUV operations; hence the BBO is employed to find feasible and near optimal solutions in competitive CPU time. More importantly, the main contribution of this paper is the proposed reactive hybrid structure which reduces run time by splitting the operation terrain to smaller zones for the path planner and also its comprehensive application for both mission management and path planning regardless of the employed algorithm, which has not been addressed before. The proposed autonomous/reactive system is implemented in MATLAB®2016 and its performance is statistically analyzed.

## 2    Mathematical Representation of the Underwater Terrain

Existence of prior information about the terrain, location of coasts and static obstacles as the forbidden zones for deployment, position of the start, target, and waypoints in operating area promotes AUV's capability in robust path planning. To model a realistic marine environment, a three dimensional terrain in scale of $\{10\times10\text{ km}(x\text{-}y), 1000\text{ m}(z)\}$ is considered based on realistic example map presented by Fig.1, in which the operating field is covered by uncertain static-moving objects, several fixed waypoints and variable ocean current. A k-means clustering method is employed to classify the coast, water and uncertain area of the map. To this purpose, a large map with the size of 1000×1000 pixels is conducted that presents 10 km square area for the mission planner and a smaller part of the map in size of 350×350 pixels that corresponds to 3.5 km square is selected to implement and test the local path planner's performance. In this map each pixel corresponds to 10 m square space. On the other hand, the operation environment is modeled by geometrical network. Therefore, nodes in the network present waypoints in operation area in which various tasks assigned to passible distance between connected nodes in advance. Hence, every edge has weight and cost that is combination of tasks priority, tasks completion time, length of the connection edges and the time required for traversing the edges. The waypoints location are randomized once in advance according to ~U(0,10000) for $P^i_{x,y}$ and ~U(0,1000) for $P^i_z$ in the joint water covered sections of the map. The clustered sections in both small and large scale maps are presented in Fig.1.

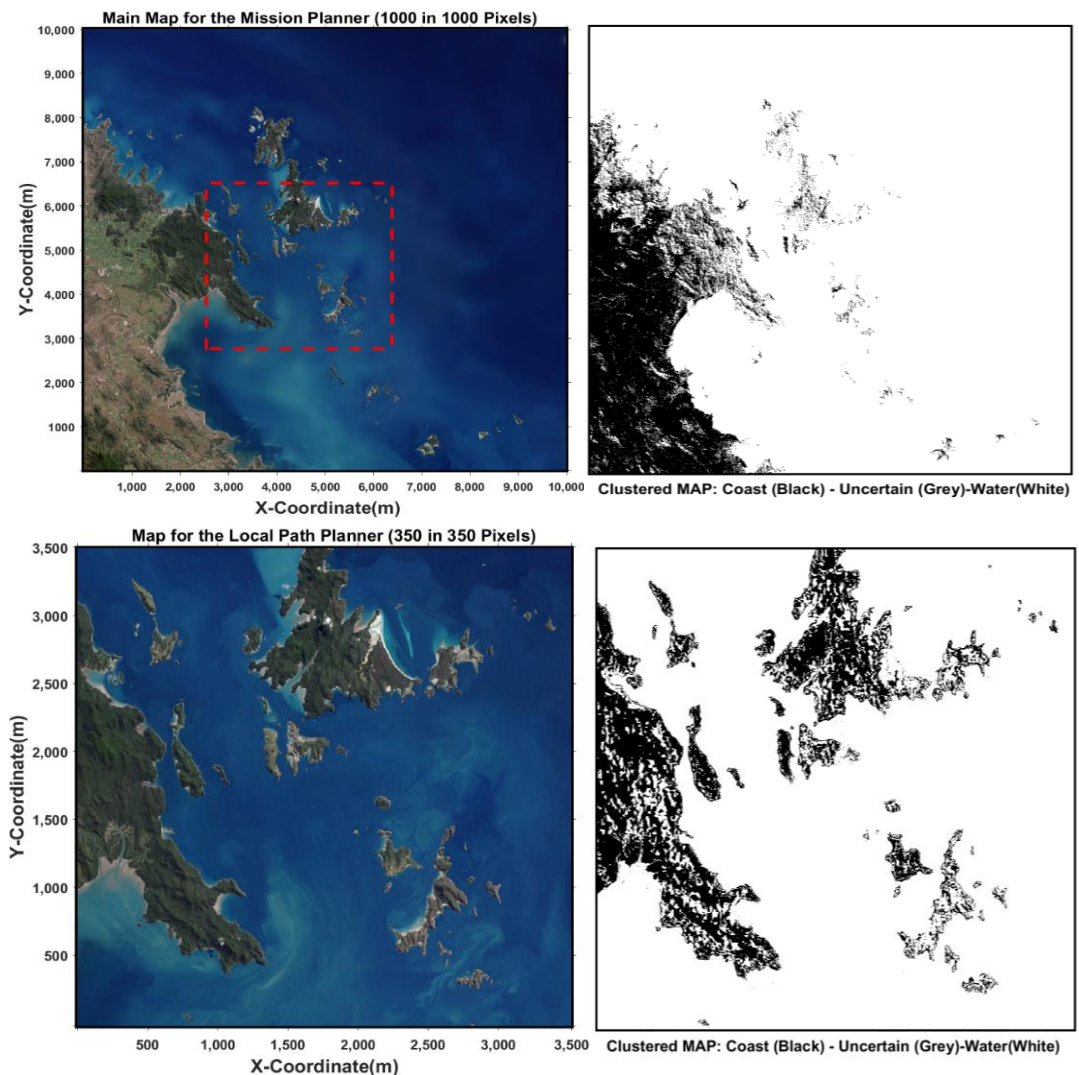

**Fig.1.**    The original and clustered map for both mission planner and the local path planner. In the clustered map

The clustered map is converted to a matrix format, in which the matrix size is same to map's pixel density. The corresponding matrix is filled with value of zero for coastal sections (forbidden area in black color), value of (0,0.35] for uncertain sections (risky area in grey color), and value of one for water covered area as valid zone for vehicles motion that presented by white color on the clustered map. The utilized clustering method is capable of clustering any alternative map very efficiently in a way that the blue sections sensed as water covered zones and other sections depending on their color range categorized as forbidden and uncertain zones.

In addition to offline map, three types of static uncertain static obstacles, afloat and self-motivated moving objects are considered in this research to cover different possibilities of the real world situations. Movement of afloat objects are considered to be affected by current force, where the self-motivated moving obstacles are considered to have a motivated velocity in additional to current effects that shift them from arbitrary position A to a random direction. Obstacle's velocity vectors and coordinates can be measured by the sonar sensors with a specific uncertainty that modelled with a Gaussian distribution; hence, each obstacle is presentable by its position ($\Theta_p$), dimension (diameter $\Theta_r$) and uncertainty ratio. Position of the obstacles initialized using normal distribution of $\mathcal{N}\sim(0,\sigma^2)$ bounded to position of two targeted waypoints by the path planner (e.g. $P^a_{x,y,z} < \Theta_p < P^b_{x,y,z}$), where $\sigma^2 \approx \Theta_r$. Further detail on modeling of different obstacles can be find at [27,28].

Beside the uncertainty of operating field, which has been taken into account in the modelling the operation terrain, water current can also have a detrimental effect on mission objectives. These issues, therefore, need to be address thoroughly in accordance with the type and range of the mission. The ocean current information can be captured from remote observations or can be obtained from numerical estimation models. Usually motion of the AUV is considered on the horizontal plane, because vertical motions in ocean structure are generally negligible due to large horizontal scales comparing to vertical [29]. For the purpose of this research, a 3D turbulent time-varying current has been applied that is generated using a multiple layered 2D current maps, in which the current circulation patterns gradually change with depth. The current map gets updated by applying each of the corresponding layers in a specific time step. The current map in this research modelled by:

$$V_c : \begin{cases} u_c(\vec{S}) = \left( -\Im(y-y_0) + \Im e^{\frac{-(\vec{S}-\vec{S}^O)^2}{\ell^2}} \right) \Big/ 2\pi(\vec{S}-\vec{S}^O)^2 \\ v_c(\vec{S}) = \left( \Im(x-x_0) - \Im e^{\frac{-(\vec{S}-\vec{S}^O)^2}{\ell^2}} \right) \Big/ 2\pi(\vec{S}-\vec{S}^O)^2 \\ w_c(\vec{S}) = \gamma \Im e^{\frac{-(\vec{S}-\vec{S}^O)^T}{2\lambda_w(\vec{S}-\vec{S}^O)}} \Big/ \sqrt{\det(2\pi\lambda_w)} \end{cases}$$

$$\lambda_w = \begin{bmatrix} \ell & 0 \\ 0 & \ell \end{bmatrix}; \quad X^{wc} \sim N(S^O, \lambda_w)$$

$$V_c : (u_c, v_c, w_c) = f(\vec{S}^O, \Im, \ell)$$

(1)

In (1), the $S$ represents a 2D space in $x$-$y$ coordinates, $S^o$ corresponds to the center of the turbulent, strength of the turbulent is presented by $\Im$, its radius is shown by $\ell$, and $\lambda_w$ is a covariance matrix based on radius of the turbulent. The $\gamma$ is a parameter to scale the vertical profile of the component ($w_c$) from the horizontal components ($u_c, v_c$). For generating a dynamic ocean current $V_C = (u_c, v_c, w_c)$, Gaussian noise can be applied on $\ell$, $\Im$, and $S^o$ parameters recursively, therefore we will have:

$$S^o_t = A_1 S^o_{t-1} + A_2 X^{S_x}_{(t-1)} + A_3 X^{S_y}_{(t-1)}$$

$$\ell(t) = A_1 \ell(t-1) + A_2 X^{\ell}_{(t-1)}$$

$$\Im(t) = A_1 \Im(t-1) + A_2 X^{\Im}_{(t-1)}$$

$$A_1 = \begin{bmatrix} 1 & 0 \\ 0 & 1 \end{bmatrix}, A_2 = \begin{bmatrix} U^C_R(t) \\ 0 \end{bmatrix}, A_3 = \begin{bmatrix} 0 \\ U^C_R(t) \end{bmatrix}$$

(2)

where, the current update rate in (2) is presented by $U^C_R(t)$ and $X^{S_x}_{(t-1)} \sim \mathcal{N}(0, \sigma_{S_x})$, $X^{S_y}_{(t-1)} \sim \mathcal{N}(0, \sigma_{S_y})$, $X^{\ell}_{(t-1)} \sim \mathcal{N}(0, \sigma_\ell)$, $X^{\Im}_{(t-1)} \sim \mathcal{N}(0, \sigma_\Im)$ are Gaussian normal distributions.

## 3 Application of the BBO Algorithm on Mission Planning and Local Path Planning

### 3.1 Overview of the BBO Algorithm

The BBO is an evolutionary algorithm inspired by equilibrium theory of island biogeography concept [32] that uses the idea of emigration, immigration, and total number of species in an island. The initial population of candidate solutions are coded with geographically isolated islands are known as habitats. Each habitat (solution) has a quantitative performance index corresponding to its fitness and called habitat suitability index (HSI). Respectively, the high quality habitats have higher HSI with more suitable habitation and share their useful information with low HIS habitats. Habitability depends on qualitative factors called Suitability Index Variables (SIVs) that is a random vector of integers. Accordingly, each habitat or candidate solution of $h_i$ possess a SIV design variable, immigration rate ($\lambda$), and emigration rate ($\mu$). The HSI value of each candidate solution corresponds to its fitness that should be maximized iteratively by the algorithm based on habitat's emigrating and immigrating features. The habitats population should be validated by defined objective function before starting the optimization process. Poor habitats have lower rate of emigration ($\mu$) and higher rate immigration ($\lambda$). The SIV value of the solution $h_i$ gets probabilistically modified based on habitats immigration rate. One of the habitats is probabilistically selected to transfer its SIV to an arbitrary solution $h_i$ according to $\mu$ rate of the other solutions that is known as migration process in BBO. Thereupon, mutation operation is applied as a helpful procedure to preserve diversity of the population and to steer the solutions toward the global optima. Each habitat is modified based on probability of the existence of $S$ species at time $t$ in habitat $h_i$. A habitat $h_i$ includes $S$ species at time $(t+\Delta t)$ by holding one of the following conditions:

[1] No emigration or immigration occurred: $h_i$ includes $S$ specie at $t \Longrightarrow h_i$ includes $S$ specie at $(t+\Delta t)$;
[2] Immigration is occurred: $h_i$ includes $S$ specie at $t \Longrightarrow h_i$ includes $(S-1)$ specie at $(t+\Delta t)$;
[3] Emigration is occurred: $h_i$ includes $S$ specie at $t \Longrightarrow h_i$ includes $(S+1)$ specie at $(t+\Delta t)$;

The probability $P_s(t+\Delta t)$ gives the change in number of species after time $\Delta t$, that is calculated by:

$$P_S(t+\Delta t) = P_S(t)(1-\lambda_S \Delta t - \mu_S \Delta t) + P_{S-1}\lambda_{S-1}\Delta t + P_{S+1}\mu_{S+1}\Delta t \qquad (3)$$

In (3), the $\lambda_s$ and $\mu_s$ represent the immigration and emigration rates when $h_i$ holds S. As habitat suitability improves, the number of its species and emigration increases, and the immigration rate decreases. Multiple immigration/emigration is not probable by assuming a very small $\Delta t$; hence, the $P_s$ is calculated as follows when $\Delta t \rightarrow 0$.

$$\dot{P}_S = \begin{cases} -(\lambda_S+\mu_S)P_S + P_{S+1}\lambda_{S+1} & S=0 \\ -(\lambda_S+\mu_S)P_S + P_{S+1}\mu_{S+1} + P_{S-1}\lambda_{S-1} & 1 \leq S \leq S_{max}-1 \\ -(\lambda_S+\mu_S)P_S + P_{S-1}\mu_{S-1} & S=S_{max} \end{cases} \qquad (4)$$

In (4), $S_{max}$ is the maximum species are collected in a habitat. Habitats with lower probability need to be mutated; thus, the mutation rate $m(S)$ for habitats has inverse relation to their probability and calculated as follows.

$$m(S) = m_{max}\left[\frac{1-P_S}{P_{max}}\right] \qquad (5)$$

In (5), the $m_{max}$ is the maximum mutation rate defined by user and $P_{max}$ is probability of the habitat with maximum number of species. There are two more control parameters of maximum immigration and emigration rates for BBO process that set by the user. Maximum emigration rate means all species are collected by a habitat. The BBO is applied in both mission planning and path planning problem as given in two following sections.

### 3.2 Biogeography-based Reactive Path Planner

Robust path planning is an important characteristic of autonomy. The path planning is a NP-hard optimization problem that aims to guide the vehicle toward a specific location encountering dynamics of the ocean and kinematics of the vehicle. An efficient path takes the minimum travel time/distance and is safe enough to satisfy collision constraints. The path planner in this research is capable of extracting feasible areas of a real map to determine the allowed spaces for deployment, where coastal area, islands, static/dynamic objects and ocean current are taken into account. Water current may have positive or disturbing effect on vehicles deployment, where an undesirable current is a disturbance itself and also it can push afloat objects across the AUV's trajectory. On the other hand, the desirable current can motivate AUVs motion toward its target; hence, adapting to water current variations is an important factor affecting optimality of the generated trajectory and considerably diminish the total AUV mission costs. The AUV is assumed to have freely deploying rigid body in 3-D space with six degrees of freedom. Six degree of freedom model of AUV comprise translational and rotational motion [30]. The state variables corresponding to NED and Body frame are shown by (6).

$$\begin{cases} \{n\} \rightarrow \eta : (X,Y,Z,\varphi,\theta,\psi) \\ \{b\} \rightarrow \upsilon : (u,v,w,p,q,r) \end{cases} \qquad (6)$$

In (6), $X$, $Y$, and $Z$ respectively, are the vehicle's North, $x$, East, $y$, Depth, $z$, position, and $\varphi,\theta,\psi$ are the Euler angles of roll, pitch, and yaw; and $\eta$ represents state of the AUV in NED frame. The $\upsilon$ is velocity vector of the vehicle in body frame in which u,v,w correspond to vehicle's directional velocities of surge, sway and heave; and p,q,r are the angular velocities. The kinematics of the AUV can be described by a set of ordinary differential equation given by (7). It is assumed the vehicle moves with constant thrust power which means the vehicle moves with constant water-referenced velocity $\upsilon$. Applying the water current components of $(u_C, v_C, w_C)$ ocean environment, the surge $(u)$, sway$(v)$, and have$(w)$ components of the vehicle's velocity is calculated as given by (8). The AUV's and ocean coordinates in $\{n\}$ and $\{b\}$ frame presented by Fig.2.

$$\begin{bmatrix} \dot{X} \\ \dot{Y} \\ \dot{Z} \end{bmatrix} = \begin{bmatrix} \cos\psi\cos\theta & -\sin\psi & \cos\psi\sin\theta \\ \sin\psi\cos\theta & \cos\psi & \sin\psi\sin\theta \\ -\sin\theta & 0 & \cos\theta \end{bmatrix} \begin{bmatrix} u \\ v \\ w \end{bmatrix} \qquad (7)$$

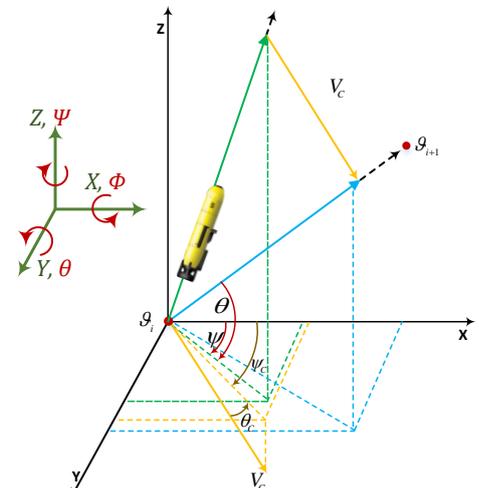

**Fig.2.** Vehicle's and ocean current coordinates in NED and body frame accordingly

$$u = |v|\cos\theta\cos\psi + |V_c|\cos\theta_c\cos\psi_c$$
$$v = |v|\cos\theta\sin\psi + |V_c|\cos\theta_c\sin\psi_c \qquad (8)$$
$$w = |v|\sin\theta + |V_c|\sin\theta_c$$

where, the $|V_C|$ is magnitude of the current speed, $\psi_c$ and $\theta_c$ give directions of current vector in horizontal and vertical planes, respectively. The path curve is generated using B-Spline method. The B-Spline curve is obtained from a sequence of control points like $\vartheta = \{\vartheta_1, \vartheta_2, ..., \vartheta_i, ..., \vartheta_n\}$ that initialized in advance in the problem search space. The mathematical description of the B-Spline coordinates is given by:

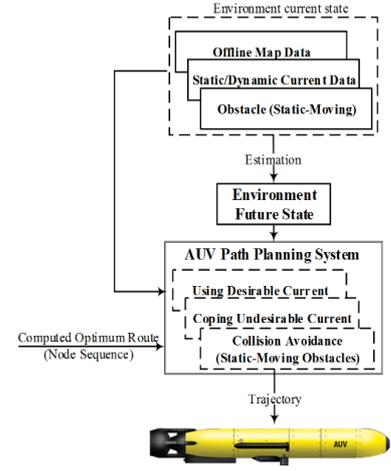

$$\left.\begin{array}{l} X = \sum_{i=1}^{n} \vartheta_{x(i)} \times B_{i,K} \\ Y = \sum_{i=1}^{n} \vartheta_{y(i)} \times B_{i,K} \\ Z = \sum_{i=1}^{n} \vartheta_{z(i)} \times B_{i,K} \end{array}\right\} \mapsto \wp_{x,y,z} = \sum_{x_s,y_s,z_s}^{|\wp|} \sqrt{\Delta X^2 + \Delta Y^2 + \Delta Z^2} \qquad (9)$$

In (9), $\wp$ is the generated path by B-Spline method, $X,Y,Z$ display the vehicle's position along the generated path. The $B_{i,K}$ is a blending function used to slice the curve, and $K$ is the order of the curve representing smoothness of the curve. For further information refer to [31]. The vehicle should be oriented along the path segments generated by (9) as given by (10).

**Fig.3**. Operation flow of the AUV path planning

$$\psi = \tan^{-1}\left(\frac{\Delta Y}{\Delta X}\right)$$
$$\theta = \tan^{-1}\left(\frac{-\Delta Z}{\sqrt{\Delta X^2 + \Delta Y^2}}\right) \qquad (10)$$

Proper adjustment of these control points plays an important role in optimality of the path curve. All control points should be located in respective search region constraint to predefined bounds of $\beta^i_\vartheta = [U^i_\vartheta, L^i_\vartheta]$. If $\vartheta_i:(x_i,y_i,z_i)$ represent one control point in Cartesian coordinates in $t^{th}$ path iteration, $L^i_\vartheta$ is the lower bound; and $U^i_\vartheta$ is the upper bound of all control points at $(x\text{-}y\text{-}z)$ coordinates. *Fig*.3 illustrates the schematic of the AUV path planning process. In the introduced BBO based path planner, habitat $h_i$ coded by the coordinates of the B-spline control points and defined as a parameter to be optimized $(P_s:(h_1,h_2,...,h_{n-1}))$, in which the habitat's corresponding HSI index is presented by a randomly initialized real vector of $n$-dimension. Each $h_i$ also involves a vector of $n$ random SIVs ($h_i:\{\chi_1, \chi_2, ..., \chi_n\}$) and approaches iteratively to its respective best position. The mechanism of the BBO based path planning is provided by Fig.4.

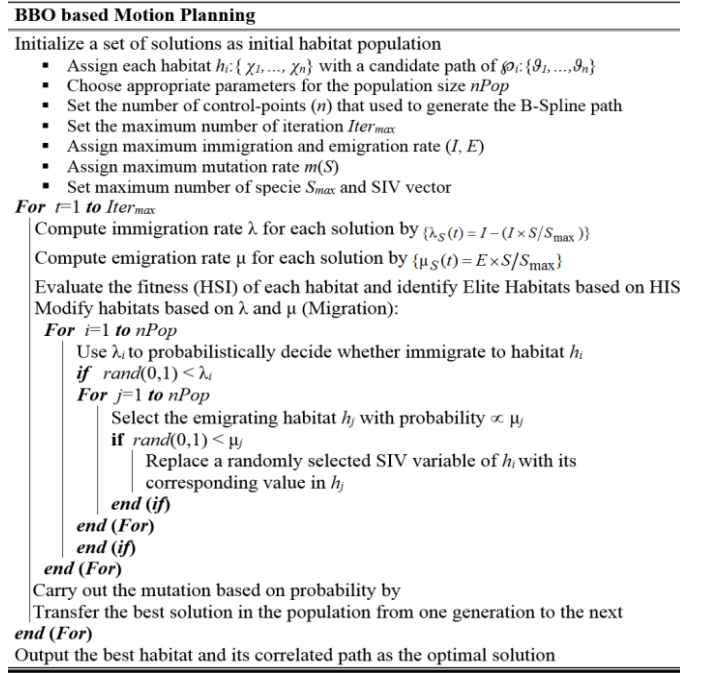

**Fig.4.** BBO optimal path planning pseudo code

### 3.2.1 Path Optimization Criterion

- **To minimize path length/time:** the path planning in this study aims to find the shortest and safest path between two waypoints taking the advantages of desirable current while avoiding collision and adverse current fields. The AUV is considered to have constant thrust power; therefore, the battery usage for a path is a constant function of the time and distance travelled. Performance of the generated path is evaluated based on overall time consumption, which is proportional to travelled distance. Assuming the constant water referenced velocity $|v|$ for the vehicle, the travel time calculated by

$$L_\wp = \sum_{x_s,y_s,z_s}^{|\wp|} \sqrt{(\vartheta_{x(i+1)} - \vartheta_{x(i)})^2 + (\vartheta_{y(i+1)} - \vartheta_{y(i)})^2 + (\vartheta_{z(i+1)} - \vartheta_{z(i)})^2}$$
$$T_\wp = \sum_{1}^{|\wp|} \frac{L_\wp}{|v|} \qquad (11)$$

In (11), the $L_\wp$ and $T_\wp$ are path length and path flight time, respectively.

- **To satisfy path constraints:** The resultant path should be safe and feasible. The environmental constraints in this paper are associated with the forbidden zones of map, collision borders of obstacles, and current disturbance. The current disturbance may cause drift to vehicle's desired motion. AUV's directional velocity components of $u,v$ and its $\psi$ orientation should be constrained to $u_{max}$, $[v_{min},v_{max}]$, and $[\psi_{min},\psi_{max}]$, while current magnitude $|V_c|$ and direction is encountered in all states along the path. The path also shouldn't cross the forbidden collision areas ($\sum M, \Theta$) associated with map or obstacles. The cost function defined as a combination path time and corresponding penalty function described in (12).

$$\wp(t) = [X(t), Y(t), Z(t), \psi(t), \theta(t), u(t), v(t), w(t)]$$

$$Cost_\wp = T_\wp + \sum_{i=1}^{n} \gamma f\left(Viol_\wp, Viol_{\sum M, \Theta}\right)$$

$$Viol_\wp = [\gamma_u, \gamma_v, \gamma_\psi] \begin{bmatrix} \max(0; u(t) - u_{\max}) \\ \max(0; |v(t)| - v_{\max}) \\ \max(0; |\psi(t)| - \psi_{\max}) \end{bmatrix} \quad (12)$$

$$Viol_{\sum M, \Theta} = \gamma_{\sum M, \Theta} \times \begin{cases} 1 & \wp_{x,y,z}(t) = Coast: Map(x, y) = 1 \\ 1 & \wp_{x,y,z}(t) \cap \bigcup_{N\Theta} \Theta(\Theta_p, \Theta_r, \Theta_{Ur}) \\ 0 & Otherwise \end{cases}$$

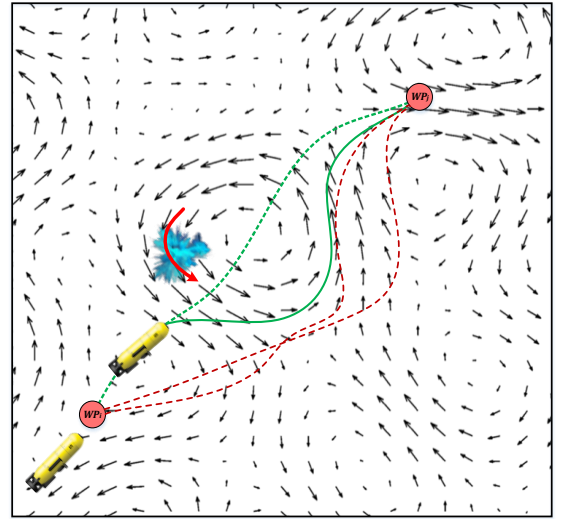

**Fig.5.** Reactive path re-planning comparing to regular path planning (Reactive re-planning reuses the subset of the past solutions)

In (12), the $\gamma f(Viol_\wp, Viol_{\sum M, \Theta})$ is a weighted violation function that respects AUV'S vehicular and collision constraints; $\gamma_u, \gamma_v, \gamma_\psi$, and $\gamma_{\sum M, \Theta}$ denote the impact of each violation in calculation of path cost.

### 3.2.2 Reactive Path Re-planning Using Previous Solution

This research applies reactive path planning as a continuous process for correcting the previous trajectory in accordance with dynamic changes of the environment. The first best path candidates are generated by applying path planning process on an initial population according to prior knowledge of the operation terrain such as offline map, initial situation of observed obstacles and initial current map information. Vehicle is obligated to take a prompt reaction to the sensed changes of the environment. To this principle, environmental situation should be simultaneously tracked and the trajectory should be refined continually. In a normal path re-planning process, the previous candidate solutions get eliminated any time that re-planning is required; then, the path planner is recalled to re-plan a new path according to new population regardless of previous solutions. Ignoring the previously found optimum path is computationally inefficient. To prevent wasting computational time and to promote the performance of the path planner, in this research a part of information from the previous solution is preserved to get reused in producing next generation of path solutions. The updated current map would contain similar features of the previous map since the currents normally change gradually over time. It is more appropriate if the previous path get refined from the current position to the destination instead of using predefined start position. The Fig.5 presents the difference between two cases addressed above.

### 3.3 Biogeography-based Adaptive Mission Planner

In the level mission planning, there should be a compromise among the mission available time, maximizing number of highest priority tasks, and guaranteeing reaching to the predefined destination, which is combination of a discrete and a continuous optimization problem at the same time (analogous to both TSP and Knapsack problem). The vehicle is requested to complete best order of the existing tasks with maximum priority in a restricted mission time. Apparently, it is impossible for a single vehicle to cover all tasks in a single mission in a large-scale operation area. Reaching to the destination is the second critical factor for the mission planner that should be taken into consideration. As mentioned earlier in section 3.1, tasks are distributed in the operating field and mapped in the feature of connected network, in which some edges of the network are assigned by tasks. The tasks should be prioritized in a way that selected edges (tasks) of the network can govern the AUV to the destination; hence, the mission planner tends to find the best collection of tasks and order them in a way that AUV gets guided to the destination while respecting the upper time threshold (mission available time) at all times during the mission. The problem involves multiple objectives that should be satisfied during the optimization process. AUV starts its mission from initial point of $P^I_{x,y,z}:(x_I, y_I, z_I)$ and should pass sufficient number of waypoints to reach on the destination at $P^D_{x,y,z}:(x_D, y_D, z_D)$. The BBO is a particular type of stochastic search algorithm that scales well with complex and multi-objective problems. The initial step in BBO is generating feasible primary solutions based on the apriori knowledge of the underwater terrain. The solutions are sequences of tasks (waypoints) that are coded by habitats in the BBO optimization process and tend to be improved iteratively according to the given optimization criterion (section 3.3.1). Each edge of the network like $q_{ij}$ includes a distance of $d_{ij}$ and a specific task of $Task_{ij}$ from a finite sequence of tasks that initialized in advance, where $Task_{ij}$ is characterized with two parameters of priority $\rho$ and absolute task completion time of $\delta$ and formulated as follows.

$$Task_{q_{ij}} : (\rho_{ij}, \delta_{ij}), \quad q_{ij} : \begin{cases} d_{ij} = \sqrt{(P_x^j - P_x^i)^2 + (P_y^j - P_y^i)^2 + (P_z^j - P_z^i)^2} \\ t_{ij} = \dfrac{d_{ij}}{|\upsilon|} + \delta_{ij} \end{cases} \quad (13)$$

In Eq.(13), $t_{ij}$ represents the required time for traversing the distance $d_{ij}$ between two waypoint of $P^i$ and $P^j$. Each edge in the graph involves the corresponding task's completion time $\delta_{ij}$ and priority $\rho_{ij}$. Tasks should be selected and ordered in a feasible feature using prior information of tasks and terrain and then get fed to the algorithm as the initial population of habitats. Following criteria assesses feasibility of a generated solutions:

- Valid solution is a sequence of nodes that commenced and ended with index of the predefined start and destination points.
- Valid solution does not include edges that are not presented in the graph.
- Valid solution does not include a specific node for multiple times, as multiple appearance of the same node implies wasting time repeating a task.
- Valid solution does not traverse an edge more than once.
- Valid solution has total completion time smaller than the maximum range of mission available time.

A priority-based strategy is conducted by this research to generate feasible solutions to the defined criteria, in which a randomly initialized priority vector is assigned to sequence of nodes in the graph. Adjacency information of the graph and provided priority vector get conducted for proper node selection. The priority vector for corresponding nodes takes positive or negative values in the specified range of [-200,100]. Afterward, nodes are added to the sequence one by one according to priority vector and graph adjacency relations. Visited nodes get a large negative priority value that prevents repeated visits them. The traversed edges of the graph get eliminated from the adjacency matrix.

```
Steps in BBO Algorithm for Mission Planning
Modelling the network terrain and transform the coordination
Generate feasible task sequences (solutions) according to priority vector and
adjacency matrix
Initialize population of habitats with generated feasible solutions
    ▪ Choose appropriate parameters for the population size nPop
    ▪ Set the maximum number of generations (iteration Iter_max)
    ▪ Assign maximum immigration and emigration rate (I, E)
    ▪ Assign maximum mutation rate m(S)
    ▪ Set S_max and SIV vector
For k=1 to Iter_max
    Compute immigration rates λ and emigration rate μ for each solution
    Evaluate the fitness (HSI) of each habitat and identify Elite Habitats based on HIS
    Modify habitats based on λ and μ (Migration):
        For i=1 to nPop
            Use λ_i to probabilistically decide whether immigrate to habitat h_i
            if rand(0,1) < λ_i
                For j=1 to nPop
                    Select the emigrating habitat h_j with probability ∝ μ_j
                    if rand(0,1) < μ_j
                        Replace a randomly selected SIV variable of h_i with its
                        corresponding value in h_j
                    end (if)
                end (For)
            end (if)
        end (For)
    Carry out the mutation based on probability
    Check the solution feasibility criterion
    Select the best solution in the population to transfer from one generation to the next
end (For)
Output the best habitat with optimal fitness value as the final best solution.
```

**Fig. 6.** BBO-based mission planning pseudo code

This process reduces the time and memory consumption for routing in large graphs. For further information about this process refer to [33]. After habitats population is initialized the algorithm start its process of finding best fitted solution according to the addressed objectives in this research (given by section 3.3.1). The whole procedure of the BBO-based mission planning is summarized in *Fig.6*.

### 3.3.1 Mission Optimization Criterion

After the habitats population is initialized, the optimization process starts to find the best fitted task sequence through the given operation graph. Maximizing the number and total priority of the selected tasks in a time interval that battery's capacity allows ($T_{Available}$) is the main goal of the mission planner. The mission time is calculated as follows:

$$T_{Mission} = \sum_{\substack{i=0 \\ j \neq i}}^{n} lq_{ij} t_{ij} = \sum_{\substack{i=0 \\ j \neq i}}^{n} lq_{ij} \left( \frac{d_{ij}}{|v|} + \delta_{ij} \right) \Rightarrow \begin{array}{l} \min(|T_{Mission} - T_{Available}|) \\ s.t. \\ \max(T_{Mission}) < T_{Available} \end{array} \quad (14)$$

In Eq.(14), the $T_{Mission}$ is the required time to complete the mission and reaching to destination. The local path planner operates in context of the mission planner and concurrently tends to generate the safe collision-free path between pairs of nodes encountering dynamicity of the terrain. The mission cost has direct relation to the required time $t_{ij}$ for passing the distance $d_{ij}$ among each pair of selected waypoints. Hence, with respect to (12) and (14), the path cost $Cost_{\wp ij}$ for any generated local path is encountered in calculation of total mission cost as follows:

$$Cost_{Mission} = \Phi_1 \left| \sum_{\substack{i=0 \\ j \neq i}}^{n} lq_{ij} (Cost_{\wp ij} + \delta_{ij}) - T_{Available} \right| + \Phi_2 \sum_{\substack{i=0 \\ j \neq i}}^{n} \left( \frac{lq_{ij}}{\rho_{ij}} \right) \quad (15)$$

$$s.t: \quad \max(T_{Mission}) < T_{Available}$$

In Eq.(15), the $\Phi_1$ and $\Phi_2$ are two positive coefficients determine amount of participation of the performance factors in the mission cost computation. After visiting each waypoint, the mission re-planning criteria should be investigated (given by section 3.3.2).

### 3.3.2 Reactive Mission Re-planning

Given a candidate sequence of waypoints and prior environmental information, the path planner provides trajectories to safely guide the vehicle to move through the waypoints until it reaches to the destination. After visiting each waypoint, the path absolute time $T_{\wp ij}$ is calculated at the end of the trajectory and compared to the expected time for passing that distance $t_{ij}$. If $T_{\wp ij}$ gets smaller than $t_{ij}$, the vehicle can continue its travel, but if $T_{\wp ij}$ exceeds the $t_{ij}$, it means unexpected challenge(s) is occurred and obviously a certain amount of the available time $T_{Avaliable}$ is taken for copping the raised situation. In such a case, the current task sequence cannot be optimum anymore; therefore, mission re-planning according to new updates would be necessary procedure. Some computational effort is devoted any time that mission re-planning is carried out; hence, this is another issue that should be taken into account. A beneficial approach for reducing the computational burden of mission re-planning is to eliminate the passed edges from the operation network (so the search space shrinks); and then change the location of the start point by the position of the current waypoint. Afterward, the mission planner tends to find a new task sequence based on new information and updated network topology. A computation cost encountered any time that mission re-planning is required. Thus, total cost for the model defined as:

$$Cost_{Total} = Cost_{Mission} + \sum_{1}^{r} T_{compute} \qquad (16)$$

In Eq.(16), the $T_{compute}$ is the time required for checking the mission re-planning criteria and computing the new optimum route, and r is the number of mission re-planning procedure. A schematic representation of the hybrid strategy is presented by a flowchart given in Fig.7.

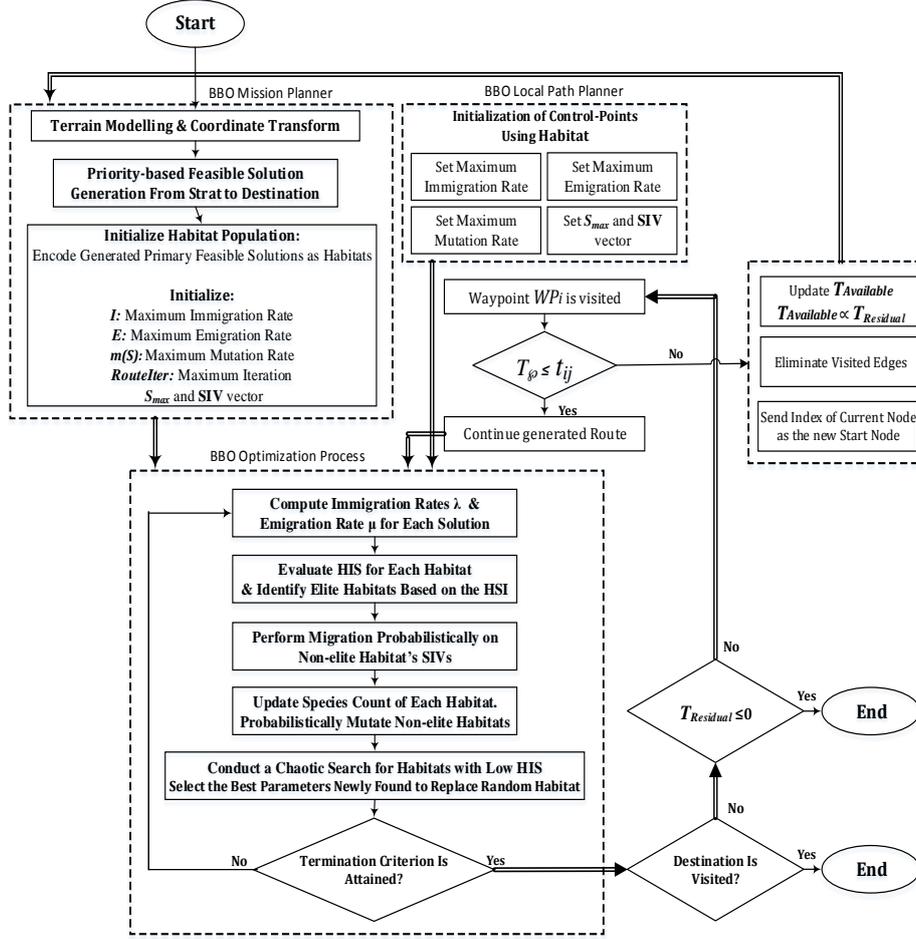

**Fig.7.** Process of the introduced BBO based hybrid mission-motion planning framework

## 4 Discussion on Simulation Results

To make the simulation process more realistic different uncertain static/dynamic obstacles are taken into account and a real map data is utilized, where the water covered area is classified by k-means method. Beside the uncertainty of operating field, water current can also affects mission objectives and sometimes makes the mission impossible. These issues, therefore, need to be address thoroughly by the designer in accordance with the type of the mission. Status of the operating field gets measured using current profiling sensors like Horizontal Acoustic Doppler Current Profiler (H-ADCP [75 kHz] and Doppler Velocity Logger (DVL). The on-board processor captures the signals and generates the real-time data of current magnitude and direction. This process is repeated until the end of the mission. Dynamic current is conducted to evaluate the path planner in this study. To this purpose, a multiple layered two dimensional current map including multiple Lamb vortices generated by Navier-Stokes equation is proposed to model the current dynamics, while the currents circulation patterns gradually change with depth. To estimate continuous circulation patterns of each subsequent layer, a recursive application of Gaussian noise is applied to the current parameters (given in section 2). As discussed earlier, the reactive path planner at lower level tends to generate the safe collision-free efficient path in a smaller scale through the waypoints produced by the mission planner; thus, the path planner operates concurrently in context of the mission planner. To evaluate the proposed hierarchical structure, the performance of the both mission and path planners are investigated as follows.

### 4.1 Simulation Results of the Reactive Local Path Planner

The path planner in this study perfectly generates optimum trajectory between two points in small scales considering dynamicity of the underwater environment. It is capable of adapting current variations, using desirable current to speed up the vehicle and save the energy, while coping with undesirable current disturbance. In all scenarios, population size for the path planner has been set on 100 habitats (candidate paths) and the algorithm runs for 100 iterations. The emigration rate is generated by $\mu = linspace(1, 0, nPop)$, and the immigration rate defined as $\lambda = 1-\mu$. The maximum mutation rate is set on 0.1. Any time that path planner is recalled by the mission planner, an offline planning system computes an initial optimum path up on available information in 100 iterations. Afterward, online re-planning system dynamically generates optimum path based on observed change in the environment to handle raised difficulty. In this research the performance of the local path planner is investigated through the different realistic scenarios:

**Scenario-1:** The first scenario includes a simple ocean environment and the path behaviour is investigated against the water current variations (shown by Fig.8).

**Scenario-2:** The second case demonstrates computing the optimum trajectory in a complex scenario in which static obstacles, afloat objects, dynamic obstacles with different degree of uncertainty, and variable spatiotemporal current field are encountered (presented in three dimensional format by Fig.9).

**Scenario-3:** This scenario as shown in Fig.10 is made even more complex by including real map data along with similar assumptions used in Scenario-2.

The proposed local path planner is able to re-generate the trajectory according to latest current updates in the vicinity of the vehicle. Fig.8 represents the behaviour of the generated B-spline path curve to current updates. In Fig.8 the colormap represents the current magnitude that is calculated according to correlations of current vectors. In this figure the lighter colors correspond to higher intensity currents. As the intensity of the current magnitude gets reduced the colour also gets darker the toward dark blue. It is clear from Fig.8 the path planner accurately copes with current changes as the red dashed line corresponeding to previouse optimum path is corrected to new path presented by thicker black line scaled from start to destination whithin three time steps. each time step shows a current update. In a dynamic path planner, when the current field update get significantly greater rate than the path update rate, the probability of trajectory failure increases respectively. So the fitness values of the produced path decreases as the rate of current update increases.

In the simulation of the second scenario, the obstacles are generated randomly from different categories and configured individually with random position by the mean variance of their uncertainty distribution and velocity proportional to current magnitude presented by Fig.9. The reactive path planner on this research simultaneously tracks the measurements of the terrain status. The trajectory should be re-planned simultaneously to avoid crossing the corresponding collision boundaries and this process repeated until vehicle reaches to the target point. Updates of the obstacles or current status regularly measured from the on-board sonar sensors and their behaviour in one step forward can be estimated according to their present state. The trajectories transformation based on 4 step current updates encountering randomly generated uncertain obstacles is presented by Fig.9, where each current update presented by a separate layer. Each current update is applied in a specific time interval that obtained from division of number of iteration to number of T-Series. Fig.10 presents the path behaviour to assumptions given by the third scenario.

Fig. 10 presents generated and replanned path in an example map of the "Whitsunday Islands", in which the terrain gets updated iteratively. The coastal areas in Fig.10 are sensed as impassable and forbidden sections for vehicles deployment. Iterations are divided into number of T-Series (which here T-Series =4) and each part is considered as a time step for updating the current. The obstacles dimension and direction changes by time in the corresponding operation window at each time step. The obstacles collision boundary grows gradually with time. The thicker black lines in Fig.10 are the new generated optimum path based on last updates of the environment.

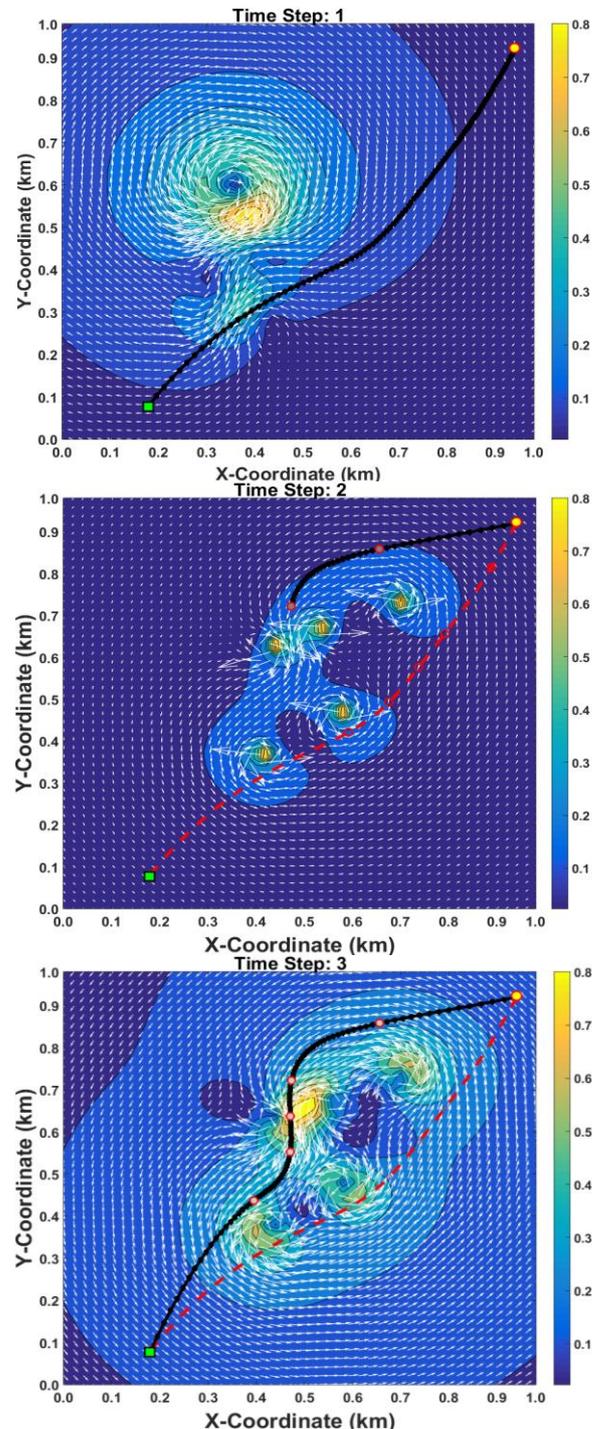

**Fig.8.** Path behaviour to current updates in three time steps where the lighter sections corresponds to higher intensity currents

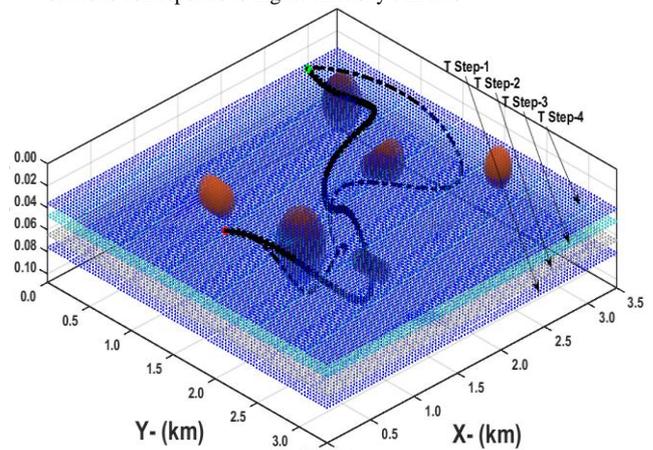

**Fig.9.** Path behaviour to current updates in four time steps encountering random combination of different type of obstacles in a 3-D space

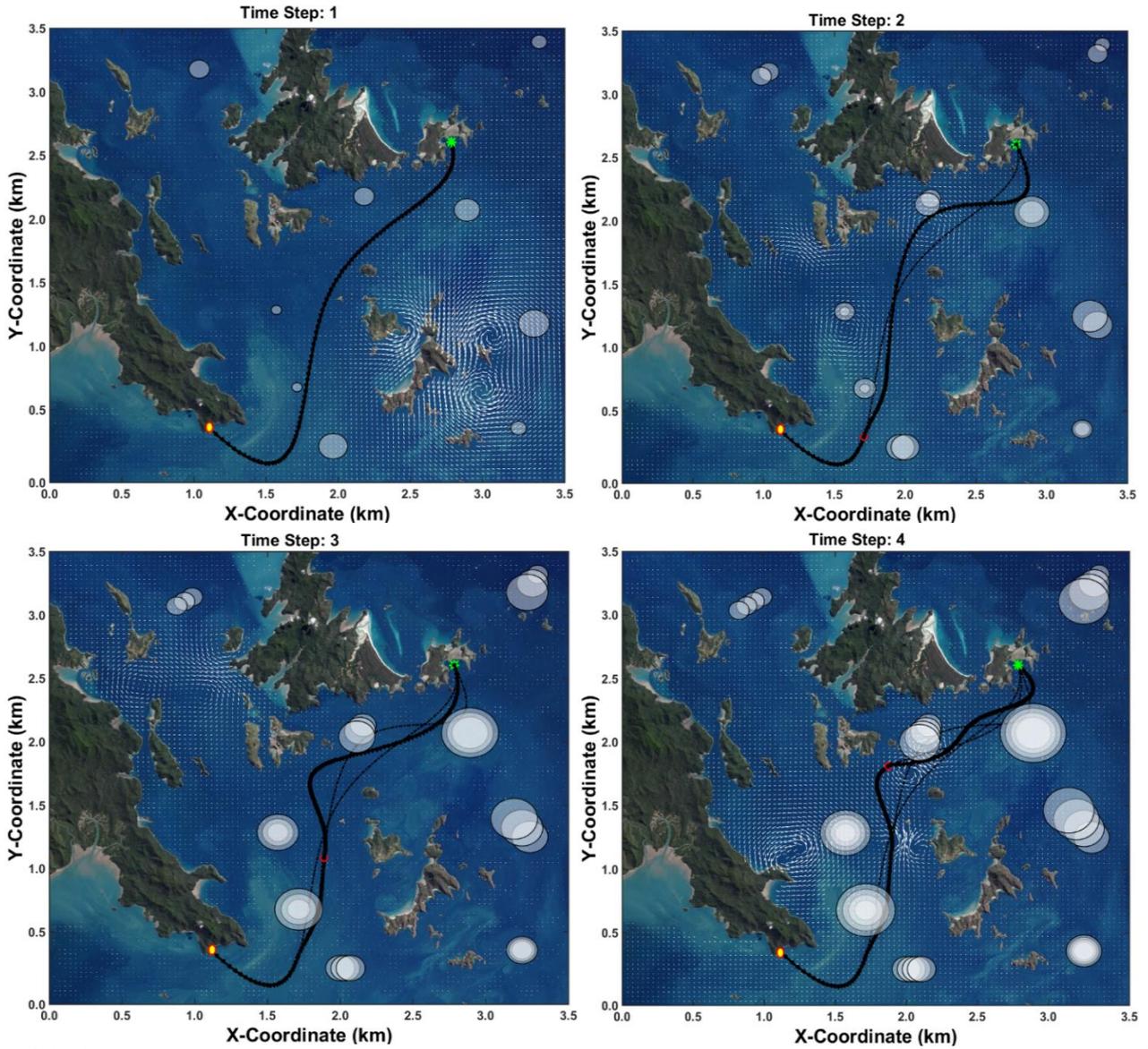

**Fig.10.** Path behaviour to update of environment encountering realmap data, different type of randomly generated obstacles and dynamic current map updated in four time steps

It is derived from simulation results in Fig.10, the proposed dynamic path re-planning methodology is capable of reusing the previous planning history to achieve more optimized path and save computation time (the offline initial path represented by thinner dashed line and regenerated online path is shown by thicker black line). The yellow and red circles in Fig.10 respectively present the start position and location of the vehicle on the map; and the green star denotes the target point. In this case the re-planned path can takes a detour to use the favourable current to speed up and save energy in its motion toward the target point.

As given by Eq.(12), the path cost is defined as function of path travel time where the vehicular and environmental constrains are taken into account. If vehicles trajectory does not cross inside the corresponding collision boundary, no collision will occur. Considering the cost variations in Fig.11(a) it is noted the algorithm experiences a moderate convergence by passing iterations. It is also noteworthy to mention from analyses the results, the cost variation range decreases in each iteration which means algorithm accurately converges to the optimum solution with minimum cost. The simulation results in Fig.11(b), shows that the proposed path planning accurately satisfies the collision and vehicular constraints as the violation for path population is diminishing iteratively which means algorithm successfully manages the path toward eliminating violations. Tracking the variation of the mean cost and mean violation in Fig.11(a,b) (presented by the black line in the middle of the error bar graphs), declares that algorithm enforces the solutions to approach the optimum answer (path) with minimum cost and efficiently manages the path to eliminate the penalty within 100 iterations. Altogether, it is derived from simulation results, the applied path planning strategy is accurate against current immediate updates. Moreover, increasing the complexity of the obstacles, increases the problems complexity, however, it is noted from simulation results in Fig.9 and Fig.10 that the algorithm is able to accurately handle the collision avoidance regardless of type or number of appeared obstacles in all three scenarios. The trajectory is re-planned simultaneously to avoid turbulent or crossing the corresponding collision boundaries and this process repeated until vehicle reaches to the target point. Further, it is outstanding in Fig.11 that the algorithm satisfies collision and vehicular constraint while minimizing the path cost that is proportional to path length and time.

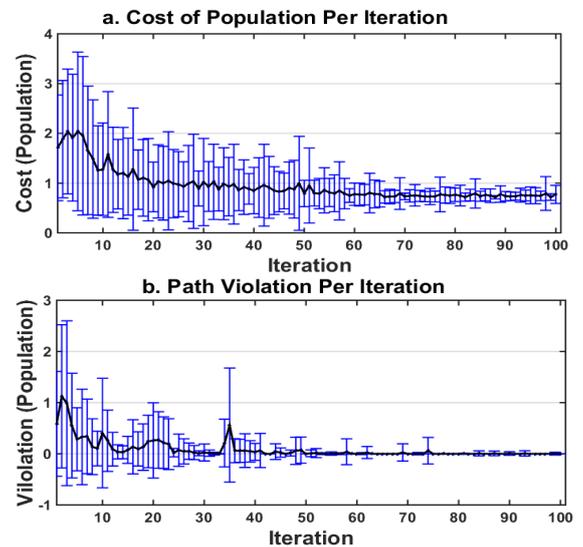

**Fig.11.** Variations of path cost and violation per iteration. Cost is function of path time and vehicular-collision violations

### 4.2 Simulation Results of the hierarchical reactive mission planning system

To evaluate the proposed hybrid reactive system, additional to local path planner that is evaluated above, the mission planner and concordance of two mission-motion planners should be investigated. To this purpose, some performance metrics have been investigated through the subsets of representative Monte Carlo simulations presented and discussed by Fig.12 to 15. The BBO, in this circumstance, is configured with habitat population size of 150, iterations of 200, and the Monte Carlo simulations is set on 100 runs, emigration rate, immigration rate , and maximum mutation rate are set on $\mu=0.2$, $\lambda=1-\mu$, and $m(S)=0.5$, respectively.

The system aims to take the maximum use of the mission available time ($T_{Available}$), increase the mission productivity by collecting best sequence of tasks fitted to $T_{Available}$, guarantee on-time termination of the mission; and concurrently ensuring the vehicles safety by copping dynamic unexpected environmental challenges during the mission. As regards, the vehicle's safe deployment is carried out by the local path planner. The stability of the model in time management is the most critical factor representing robustness of the method. To this end, the robustness of the hierarchal model in enhancement of the vehicles autonomy in terms of mission time management is evaluated by testing 100 missions' through individual experiments with the same initial condition that closely matches actual underwater mission scenarios and presented by Fig.12 to 15. The number of waypoints is fixed on 40 nodes for all Monte Carlo runs, but the topology of the graph is changed randomly based on a Gaussian distribution on the problem search space. The time threshold ($T_{Available}$) also fixed on 4 hours (14,400 seconds).

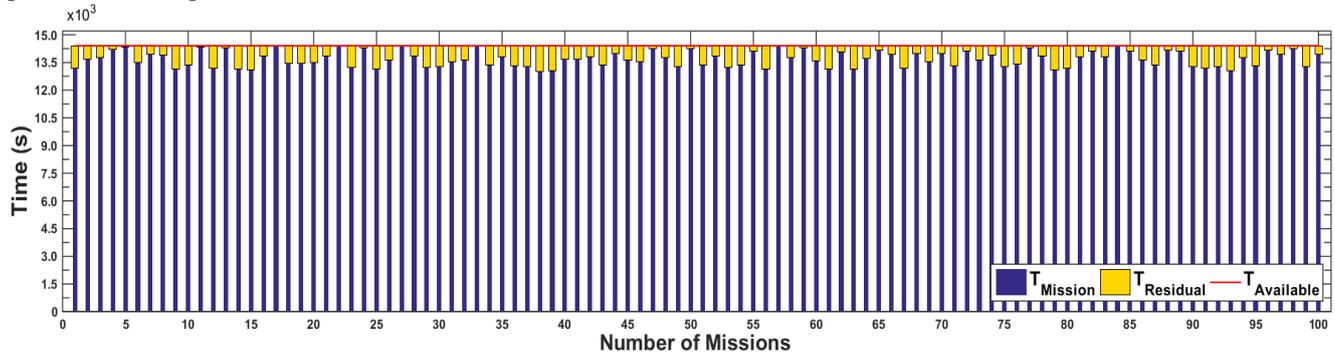

**Fig.12.** Performance of the model in maximizing the mission time constraint to available time threshold

To establish appropriate collaboration between the low and high level mission-motion planners, the correlation between the expected time for passing a specific distance ($t_{ij}$) and the generated path time ($T_\wp$) is another important performance index affecting concordance of the whole system that investigated and presented by Fig.13. The $t_{ij}$ for the local planner is determined from the estimated mission time (given by Eq.(14)).

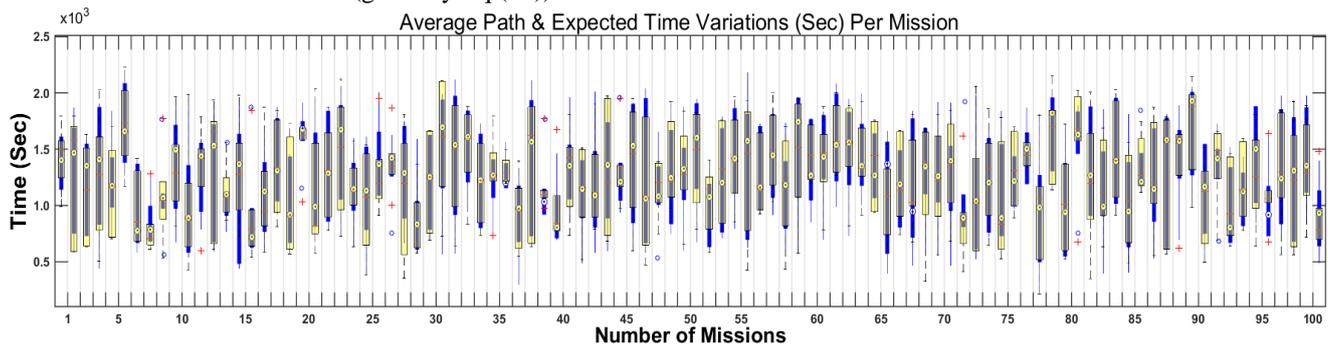

**Fig.13.** Stability of the hierarchical reactive system in managing correlation between expected and path time ($t_{ij}$ and $T_\wp$) in multiple recall of the local path planner in each mission through the 100 Monte Carlo simulations

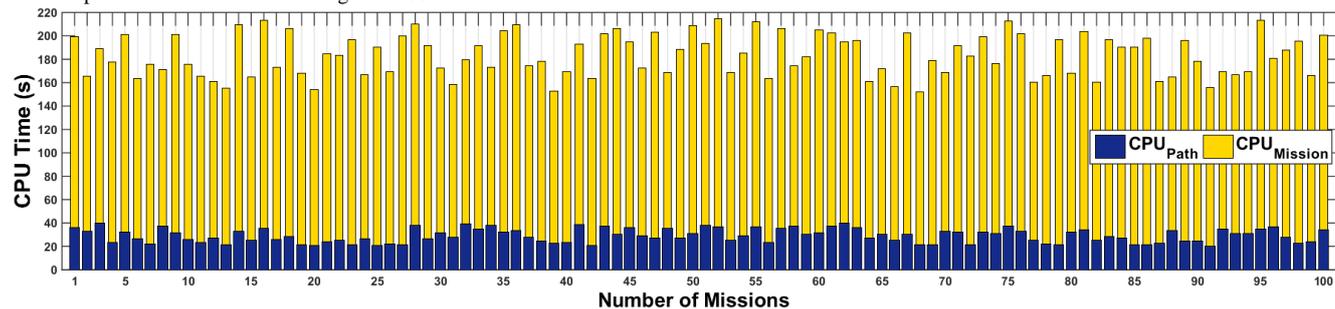

**Fig.14.** Relative proportions of CPU time used for a single run of the path planning compared to overall CPU time for mission which includes multiple operation of the path planner.

The most appropriate outcome for a mission is completion of the mission with the minimum residual time ($T_{Residual}$), which means maximizing the use of mission available time. The mission time ($T_{Mission}$) is the summation of time taken for passing through the waypoints (produced by the local path planner) and completion time for covered tasks along paths. It is clear in Fig.12 the proposed hierarchal reactive system is capable of taking maximum use of available time as the $T_{Mission}$ in all experiments approaches the $T_{Available}$ and meet the above constraints denoted by the upper bound of 14400 sec (4 hours) presented by red line. Respectively, the $T_{Residual}$ that has a linear relation to $T_{Mission}$ should be minimized but it should not be equal to zero that is accurately satisfied considering variations of $T_{Residual}$ over 100 missions. Considering the fact that reaching to the destination is a big concern for vehicles safety, a big penalty value is assigned to strictly prevent missions from taking more time than $T_{Available}$.

Existence of a reasonable correlation between $T_{\wp}$ and $t_{ij}$ values in each path planning operation is critical to total performance of the hierarchical model, which means a big difference between these two parameters cause interruption in cohesion of the whole system. It is derived from simulation result in Fig.13 that variations of both $T_{\wp}$ and $t_{ij}$ are lied in similar range and very close to each other in which the $T_{\wp}$ and $t_{ij}$ are presented with yellow transparent boxplot and blue compact boxplot, respectively.

Another highlighted factor is the computational time for mission and path planning operations. The reason for highlighting the CPU time is that the local planner must operate synchronous to the mission planner; thus, a large CPU time again causes interruption to concurrent operation of two planners, which this issue flaws the routine flow and cohesion of the whole system. Fig.14 presents the relative proportions of CPU time used for a single run of the path planning compared to overall CPU time for mission which includes multiple operation of the path planner. It is noteworthy to mention from analyze of the simulation results in Fig.14 that the model takes a very short CPU time for all experiments that makes it highly accurate for real-time application. As the local planner operates multiple time in context of the mission planner, obviously it takes shorter time and lower cost, comparatively. It is evident from Fig.14 that the range of variations of CPU time for both planners is almost in similar range in all experiments that proves the inherent stability of the model's real-time performance. The results obtained from Monte Carlo analyze (presented by Fig12 to Fig.14) demonstrate the inherent robustness and drastic efficiency of the proposed hierarchical reactive system in terms of time management and increasing mission productivity by maximizing $T_{Mission}$ while each mission is terminated before vehicle runs out of time/energy, as in all experiment $T_{Residual}$ gets nonzero positive value.

For proper visualization of the dynamic mission planning process, one of the experiments is proposed by Fig.15 in which rearrangement of order of tasks and waypoints according to updated $T_{Residual}$ is presented through three re-planning procedure. In the beginning of the mission the mission planner is recalled to generate the initial optimum task sequence fitted to available time. Referring to Fig.15, the initial sequence encapsulates number of 17 tasks (edges) with total weight of 47, and estimated completion time of $T_{Mission}$=13976 (sec). The local path planner takes the waypoint sequence and starts generating feasible path through the listed waypoints in the initial sequence; hence the local planner adopts the first pair of waypoints and generate a feasible trajectory between them with time of $T_\wp=923.2(sec)$ which is smaller than expected travel time $t_{ij}=987.7(sec)$. In cases that $T_\wp$ is smaller than $t_{ij}$, the mission replanting flag is zero - the initial sequence is still valid and optimum- so the vehicle is allowed to follow the next pair of waypoints included in initial sequence. After each run of the path planner, the path time $T_\wp$ is reduced from the $T_{Residual}$. Accordingly, waypoints are shifted to the local planner and the same process is repeated until the $T_\wp$ exceeds the $t_{ij}$. In this circumstance, the mission re-planning flag gets one, which means some of the available time is wasted in passing the distance between those nodes. Hence, the $T_{Residual}$ gets updated and the global planner is recalled to rearrange the tasks order and regenerate a new task sequence from the existing node. In simulation results presented in Fig.15, the mission planner is recalled for 3 times and the local path planner is called for 11 times. This interplay among the path planner and mission planner continues until vehicle reaches to the destination (success) or $T_{Residual}$ gets a minus value (failure: vehicle runs out of battery).

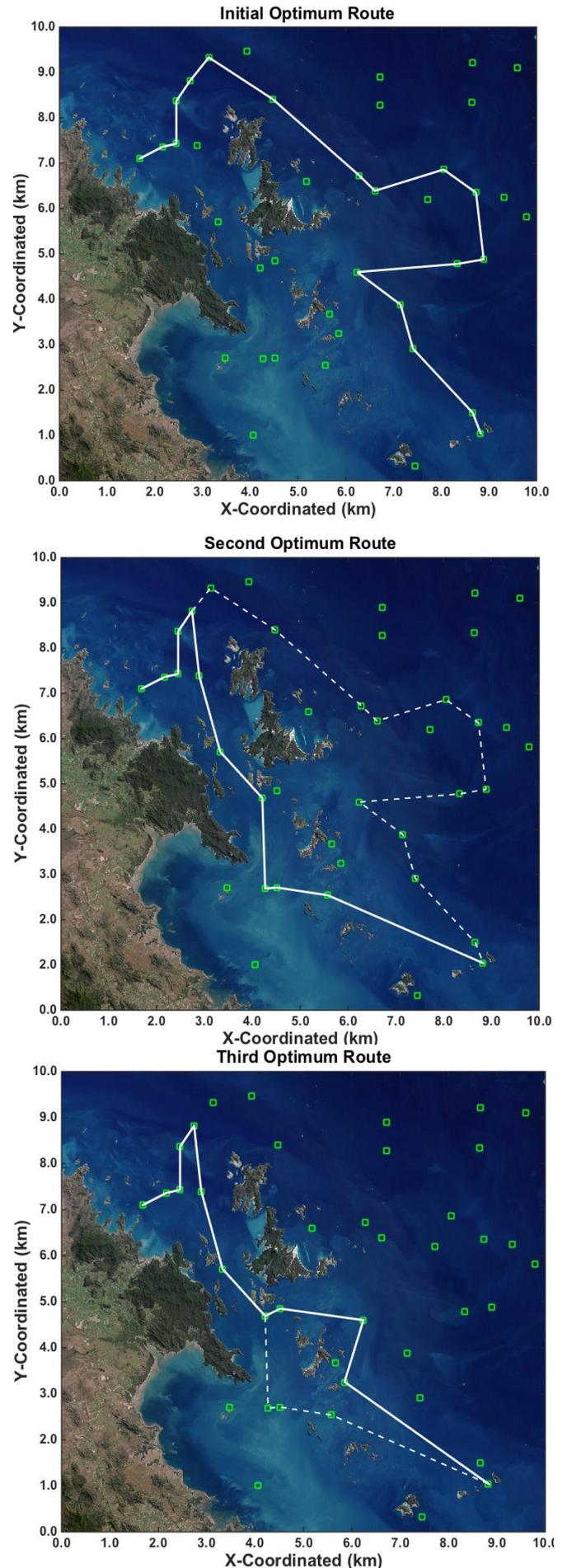

**Fig.15.** The mission planning/replanning and tasks rearrangement based on updated terrain and available time

## 5  Conclusion

A two-level dynamic mission-motion planning for AUV's optimal task-assignment and time management has been developed. The higher level mission planner provides the best sequence of waypoints (tasks) with respect to the mission objectives. By doing this, the optimal deployment zone is selected and the search area is minimized. Afterward, the local reactive path planner starts generating optimal collision-free path through the listed waypoints; copping any unexpected environmental changes is done by correcting the old path or re-generating a new path. The spatio-temporal variability of the operating field is tracked and then based on new situational awareness of surrounding environment the mission planner updates the tasks orders. The local path planner in current study mainly focuses on finding the optimal trajectory by using the advantages of desirable current field and avoiding obstacles and no-fly zones. The proposed hybrid reactive system takes advantage of the BBO algorithm on both mission and path planners where the main objectives of the system is to have a beneficial mission, manage the mission time considering total available time, task priority assignment, and safe deployment. Knowing the fact that existing approaches mainly are able to cover only a part of problem of ether motion/path planning or task assignment based on offline map, the provided framework is remarkable contribution to improve AUV's autonomy in both higher/lower level due to its comprehensive mission-motion planning capability that covers shortcomings associated with previous strategies used in this scope. Future work will concentrate on the design of a modular architecture with an additional module of predictor–corrector encompassing the ability of at least one step ahead prediction of situational awareness of the environment that results in more flexibility for successful mission in real-world experiments.